\documentclass[10pt, a4paper]{article}
\usepackage[final]{lrec2026} 

\usepackage{amsmath,amsfonts,bm}
\usepackage{microtype}
\usepackage{tikz}
\usepackage{pgfplots}
\pgfplotsset{compat=1.18}
\usetikzlibrary{matrix, positioning, calc}
\usepackage{makecell}
\usepackage{algorithm}
\usepackage{algpseudocode}
\usepackage{tabularx}
\newcolumntype{C}{>{\centering\arraybackslash}X}
\usepackage{subcaption}
\usepackage{mathtools}
\usepackage{float}
\floatstyle{plain}
\usepackage{pifont}
\usepackage{multirow}
\usepackage{tcolorbox}
\usepackage{cleveref} 
\crefname{algorithm}{alg.}{algorithms}
\Crefname{algorithm}{Algorithm}{Algorithms}

\definecolor{mygreen}{rgb}{.21, .49, .13}

\newcommand{\segmenter}{$\texttt{much\_segmenter}$}

\newfloat{algorithm}{htbp}{loa}
\floatname{algorithm}{Algorithm}

\makeatletter
\def\blfootnote{\def\Hy@Warning##1{}\gdef\@thefnmark{}\@footnotetext}
\makeatother

\title{MUCH: A Multilingual Claim Hallucination Benchmark}

\name{
\begin{tabular}{c}
Jérémie Dentan$^1$, Alexi Canesse$^1$, Davide Buscaldi$^{1,2}$,\\
Aymen Shabou$^3$, Sonia Vanier$^1$
\end{tabular}
} 

\address{$^1$LIX (École Polytechnique, IP Paris, CNRS), $^2$LIPN (Sorbonne Paris Nord), $^3$Crédit Agricole SA \\
\{jeremie.dentan, sonia.vanier\}@polytechnique.edu\\}

\abstract{Claim-level Uncertainty Quantification (UQ) is a promising approach to mitigate the lack of reliability in Large Language Models (LLMs). We introduce MUCH, the first claim-level UQ benchmark designed for fair and reproducible evaluation of future methods under realistic conditions. It includes 4,873 samples across four European languages (English, French, Spanish, and German) and four instruction-tuned open-weight LLMs. Unlike prior claim-level benchmarks, we release 24 generation logits per token, facilitating the development of future white-box methods without re-generating data. Moreover, in contrast to previous benchmarks that rely on manual or LLM-based segmentation, we propose a new deterministic algorithm capable of segmenting claims using as little as 0.2\% of the LLM generation time. This makes our segmentation approach suitable for real-time monitoring of LLM outputs, ensuring that MUCH evaluates UQ methods under realistic deployment constraints. Finally, our evaluations show that current methods still have substantial room for improvement in both performance and efficiency.
\\ \newline 
\begin{center}
  \begin{minipage}{0.33\linewidth}
    \centering
    \raisebox{-0.2\height}{\includegraphics[width=1em]{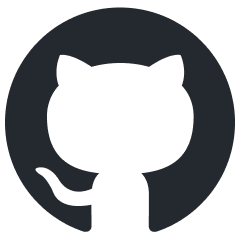}}%
    \hspace{0.5em}%
    {\small\texttt{\href{https://github.com/orailix/much}{\tt orailix/much}}}
  \end{minipage}
  \begin{minipage}{0.33\linewidth}
    \centering
    \raisebox{-0.2\height}{\includegraphics[width=1em]{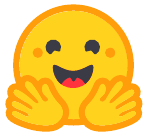}}%
    \hspace{0.5em}%
    {\small\texttt{\href{https://huggingface.co/datasets/orailix/MUCH}{\tt orailix/MUCH}}}
  \end{minipage}
  \begin{minipage}{0.33\linewidth}
    \centering
    \raisebox{-0.2\height}{\includegraphics[width=1em]{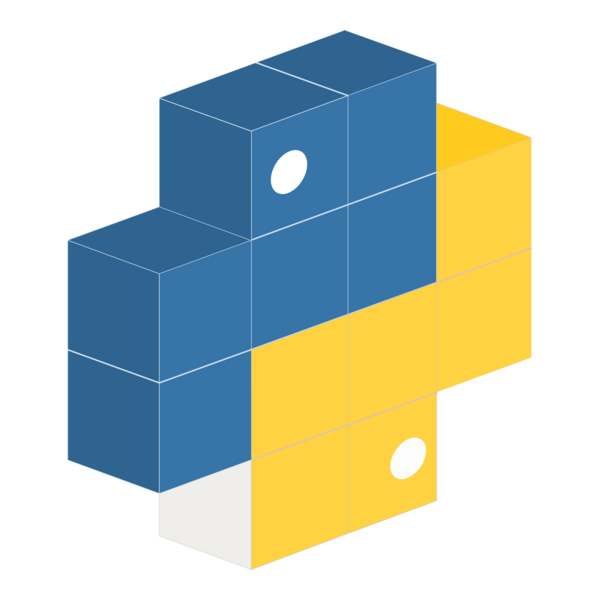}}%
    \hspace{0.5em}%
    {\small\texttt{\href{https://pypi.org/project/much-segmenter/}{\tt much-segmenter}}}
  \end{minipage}
\end{center}}

\begin{document}

\maketitleabstract

\pagestyle{plain}
\thispagestyle{plain} 
\pagenumbering{arabic}

\section{Introduction} \label{sec:intro}

\blfootnote{To appear in Proceedings of LREC 2026.}

Despite significant improvements in their performance, the reliability of Large Language Models (LLMs) remains an open challenge, as these models are prone to producing plausible yet non-factual content, usually referred to as hallucinations~\citep{huang_survey_2025, zhang_sirens_2025, sahoo_comprehensive_2024, ji_survey_2023}. To mitigate the consequences of factuality hallucinations, Uncertainty Quantification (UQ) techniques have been developed to estimate an LLM’s confidence in its responses~\citep{shorinwa_survey_2026, xia_survey_2025, geng_survey_2024, huang_survey_2024}. Most UQ benchmarks focus on a single aggregate uncertainty score for the entire generation. However, this response-level score does not provide a fine-grained view of uncertainty, and a long LLM generations risk being rejected even if only a small part is incorrect. To address these limitations, claim-level UQ methods were recently proposed~\citep{fadeeva_fact-checking_2024}. These methods provide a local uncertainty score for each distinct idea in the output.

\newcolumntype{Y}[1]{>{\raggedright\arraybackslash}m{#1}}
{
\setlength{\tabcolsep}{2pt}
\begin{table}[t!]
	\centering
	\begin{tabularx}{0.48\textwidth}{Y{2.2cm} Y{1.0cm}Y{1.9cm}Y{0.7cm}c}
		\Xhline{.8pt}
		\noalign{\vskip .75mm}
		\textbf{Benchmark} & \textbf{Scale} & \textbf{Segmenter}    & \textbf{Size} & \textbf{Logits} \\
		\Xhline{.8pt}
		\noalign{\vskip .75mm}
		Mu-SHROOM           & Span & Human                 & 2.4k          & \color{red!75!black}\ding{55}       \\
		LM-Polygraph         & Claim & LLM                  & 818           & \color{red!75!black}\ding{55}       \\
		PsiloQA              & Span & LLM                   & 70k           & \color{red!75!black}\ding{55}       \\
		HalluEntity          & Entity & Human+LLM           & 157           & \color{red!75!black}\ding{55}       \\
		RAGTruthQA           & Span & Human                 & 18k           & \color{red!75!black}\ding{55}       \\
		FAVA                 & Span & Human                 & 902           & \color{red!75!black}\ding{55}       \\
		\textbf{MUCH} & Claim & Algorithmic          & 4.8k & \color{green!50!black}\checkmark     \\
		\Xhline{.8pt}
	\end{tabularx}
	\caption{Existing benchmarks: Mu-SHROOM~\protect\citeplanguageresource{mushroom}, LM-Polygraph~\citep{vashurin_benchmarking_2025}, PsiloQA~\citep{rykov_when_2025}, HalluEntity~\citep{yeh_halluentity_2025}, RAGTruthQA~\citep{niu_ragtruth_2024}, FAVA~\citep{min_factscore_2023}, and MUCH (ours). We show the hallucination scale, segmentation method ``Segmenter''), dataset size, and whether logits are available.}
	\label{tab:teaser}
\end{table}
}
\begin{figure*}[!t]
\begin{center}
    \includegraphics[width=\textwidth]{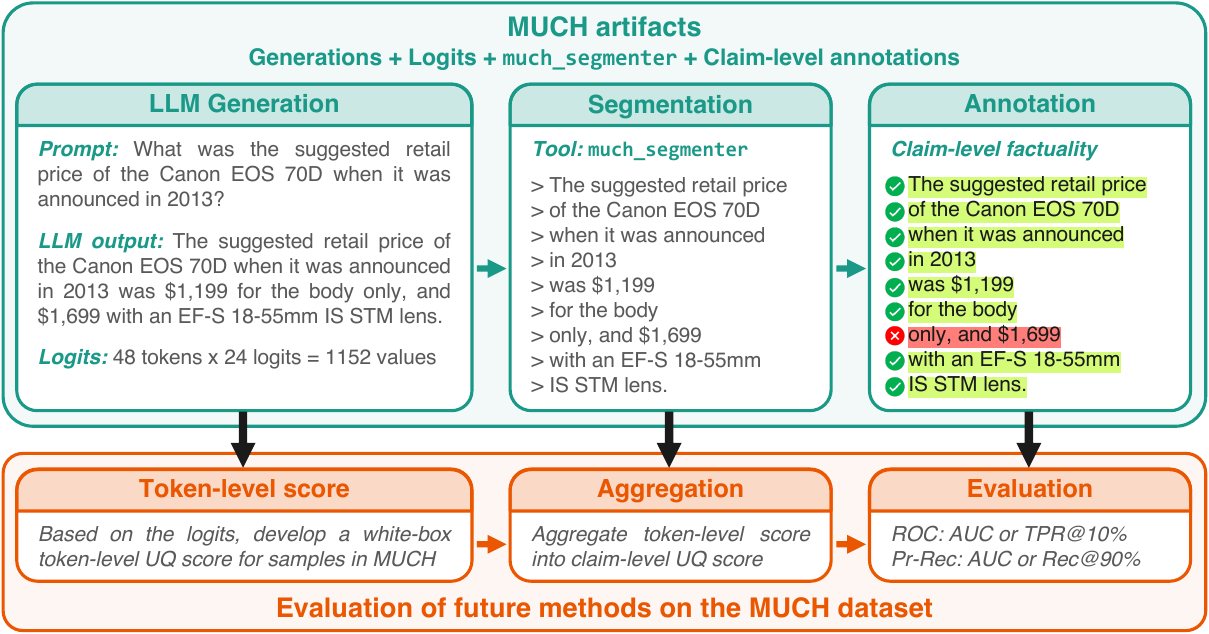}
    \caption{We open-source four artifacts as part of the MUCH benchmark: (1) 4,873 LLM generations spanning four languages (English, French, Spanish, and German) and four models (Llama 3.1 8B, Llama 3.2 3B, Ministral 8B, and Gemma 3 4B); (2) 24 logits per generation token; (3) \segmenter, a fast and reproducible claim segmenter; and (4) claim-level factuality annotations for every sample, totaling 20,751 binary annotations. This framework facilitates the evaluation of future methods, which only requires defining a new token-level score, aggregating it, and comparing it to the claim-level annotations.}
    \label{fig:teaser}
\end{center}
\end{figure*}

\paragraph{Limitations of existing UQ benchmarks}

Despite promising impact across various application scenarios, claim-level UQ remains an emerging field. We identify two key limitations in existing benchmarks, summarized in~\Cref{tab:teaser}. First, no benchmark releases multiple logits per LLM-generated token, making these datasets unsuitable for developing and evaluating new white-box UQ methods, which typically require access to multiple logits. Second, segmentation is performed under unrealistic conditions, which does not allow for a reliable estimation of UQ performance for real-time monitoring of LLM outputs in production. Existing benchmarks either split LLM generations into claims that are then annotated (``Claim'' in Tab.~\ref{tab:teaser}), or directly annotate spans within generations (``Span''), or annotate entities from human-defined claims (``Entity''). Human-based approaches are clearly impractical in production, while LLM-based ones are computationally expensive, requiring as much computation as the output generation itself. They are also stochastic and non-reproducible, limiting the generalizability of evaluation results. These limitations often force new methods to regenerate evaluation data and re-implement baselines for comparison \citep{fadeeva_fact-checking_2024, farquhar_detecting_2024}

\paragraph{Benchmarking future methods with MUCH}

To address these limitations, and to support the development of new UQ approaches, we introduce MUCH, a new multilingual claim hallucination dataset and evaluation protocol, illustrated in~\Cref{fig:teaser}. The dataset contains 4,873 pairs of question and LLM-response. Importantly, we provide 24 logits per LLM-generated token, enabling future research to directly evaluate new white-box, logit-based approaches without re-generating data. We also release~\segmenter, a new claim segmentation algorithm that addresses the limitations of LLM-based and human-based segmentation. Relying on keyword and punctuation cues, it is fully deterministic and extremely fast: segmenting the entire dataset requires only 0.2\% of the computation time needed for LLM generation. It is independent of any UQ method, ensuring fair comparisons with future approaches and eliminating the need for re-annotation. Finally, we provide an annotation in $\{-1; +1\}$ to assess the factuality of each of the 20,751 claims in the dataset.

The construction of MUCH is presented in~\Cref{fig:pipeline}. The questions are taken from the Mu-SHROOM test set~\citeplanguageresource{mushroom} and consist of factual questions answerable from a single Wikipedia page provided with each question. This ensures that the factuality of responses can be evaluated objectively. We retained four European languages (English, French, Spanish and German), with approximately 200 questions per language. For each question, we generated 8 LLM responses using two different temperatures and four instruction-tuned open-weight models (Llama 3.1 8B Instruct and 3.2 3B Instruct~\citep{grattafiori_llama_2024}, Ministral 8B Insruct~\citep{mistral_ai_team_ministral_2025}, and Gemma 3 4B Instruct~\citep{team_gemma_2025}). Then we automatically annotated claim-level factuality using the corresponding Wikipedia page and two different LLMs (GPT-4o and GPT-4.1). To ensure annotation quality, we retained only the 4.8k generations for which GPT-4o and GPT-4.1 labels perfectly matched. Finally, a small portion of the dataset was double-annotated by humans, showing that the gap between automatic and human labels is comparable to inter-human variability.

\paragraph{Summary of contributions}

\begin{itemize}
    \item We release MUCH, a benchmark for claim-level uncertainty quantification in English, French, Spanish and German;
    \item We release generation logits to support the development of new white-box UQ methods;
    \item We release \segmenter, a fast and deterministic claim segmentation method;
    \item We document the construction of MUCH and compare automatic annotations to human labels to ensure quality and reproducibility;
    \item We benchmark the current best claim-level UQ methods and discuss directions for improving both performance and efficiency.
\end{itemize}

\section{Background and Related Works} \label{sec:back_uq}

\paragraph{Fact uncertainty quantification benchmarks}

Literature distinguishes between \textit{faithfulness} and \textit{factuality} hallucinations \citep{huang_survey_2025}, or alternatively between \textit{input-conflicting}, \textit{context-conflicting} and \textit{fact-conflicting} hallucinations~\citep{zhang_sirens_2025}. Our benchmark focuses exclusively on factuality hallucinations in LLM responses. Numerous response-level benchmarks have been proposed for this task, including BioASQ~\citep{krithara_bioasq-qa_2023}, TruthfulQA~\citep{lin_truthfulqa_2022}, NQ~\citep{kwiatkowski_natural_2019}, SQuAD~\citep{rajpurkar_know_2018}, SVAMP~\citep{patel_are_2021} or TriviaQA~\citep{joshi_triviaqa_2017}. However, this paper only focuses on claim-level UQ, which provides a finer-grained view of uncertainty in LLM responses than response-level approaches. The existing claim-level UQ benchmarks we are aware of are listed in~\Cref{tab:teaser}. As discussed in the Introduction, the lack of efficient and deterministic segmentation methods, along with the absence of logits, makes these benchmarks inadequate for evaluating future claim-level UQ methods in settings that generalize to real-world scenarios.

\paragraph{White-Box, Sample-Specific UQ}

UQ literature distinguishes between \textit{white-box} approaches, such as~\citet{fadeeva_fact-checking_2024, sriramanan_llm-check_2024, chen_inside_2024, kadavath_language_2022, duan_shifting_2024}, which require access to logits, internal activations, or the model itself, and \textit{black-box} approaches, such as~\citet{kuhn_semantic_2023, nikitin_kernel_2024, lin_generating_2023}, which rely only on the generated text. UQ methods can also be classified as sample-specific or population-level~\citep{sriramanan_llm-check_2024}. The former assign a UQ score to a single LLM response, while the latter assign a score to multiple LLM generations for the same prompt.

Our benchmark focuses on white-box, sample-specific UQ. First, we posit that the reliability of an LLM's output is primarily the responsibility of the provider who generates them and, consequently, has white-box access to the model. Second, substantial efforts are made to reduce inference time and computational costs. Population-level approaches require multiple generations, which contradicts these goals and limits their adoption.

\paragraph{Claim segmentation} 

All claim-level UQ methods we are aware of rely on prompt-based segmentation by an LLM or human-based segmentation (see~\Cref{tab:teaser}). However, such LLM-based segmentation is impractical in realistic scenarios for three main reasons, which substantially limits the utility of the benchmarks proposed in these works. First, LLM-based segmentation is intrinsically non-deterministic, and some papers do not report the exact LLM version used, preventing results from generalizing to future applications. Second, LLM segmentation is costly, requiring the segmenter-LLM to re-generate all claims, which is at least as costly as generating the target LLM outputs for which UQ is computed. Finally, LLM-based segmentation requires mapping target LLM output tokens to the claims produced by the segmenter-LLM, which is a non-trivial problem. For instance,~\citet{fadeeva_fact-checking_2024} report that about 5\% of claims could not be mapped to original tokens, which is unacceptable in many applications. On the opposite, our segmenter is deterministic, extremely fast (only 0.2\% of the computation cost of LLM generation), and directly aligns with the target generation tokens, making it suitable for realistic applications.

\paragraph{Existing sample-specific, claim-level methods}

Few approaches have been developed specifically for claim-level UQ. In~\Cref{baselines}, we evaluate on MUCH several baselines often cited as top-performing in existing benchmarks: CCP~\citep{fadeeva_fact-checking_2024}, SAR~\citep{duan_shifting_2024}, Token Likelihood~\citep{guerreiro_looking_2023}, Token Entropy~\citep{malinin_uncertainty_2021}, and Maximum Likelihood~\citep{aichberger_rethinking_2024}. Although some of these methods were not designed for claim-level UQ, they produce token-level scores that can be aggregated at the claim level. We did not include FOCUS~\citep{zhang_enhancing_2023}, despite its strong performance on several benchmarks~\citep{yeh_halluentity_2025, rykov_when_2025}, because it relies on attention scores, which are unavailable in modern implementations such as Flash-Attention~\citep{dao_flashattention_2022, dao_flashattention-2_2023}, making it impractical for production.
\section{Methodology} \label{sec:methodology}

\begin{figure*}[!t]
\begin{center}
    \includegraphics[width=\textwidth]{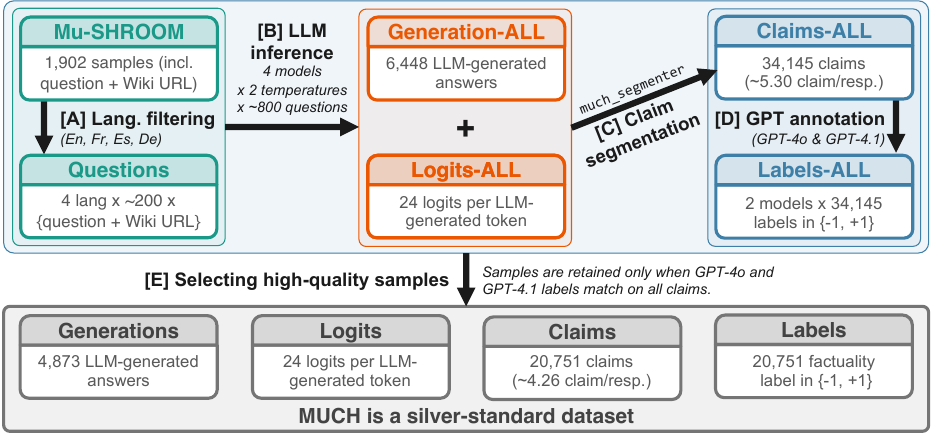}
    \caption{Construction pipeline of MUCH benchmark. We filter English, French, Spanish, and German questions from the Mu-SHROOM test set \protect\citeplanguageresource{mushroom} (see [A]). We then generate eight LLM answers per question, and retain 24 logits per generated token (see [B]). Next, we use~\segmenter~to parse LLM generations (see [C]). We automatically assign two binary labels to each claim, one using GPT-4o and one using GPT-4.1 (see [D]). Finally, we retain only high-quality annotations by filtering out samples where GPT-4o and GPT-4.1 labels mismatch on at least one claim (see [E]).}
    \label{fig:pipeline}
\end{center}
\end{figure*}

The construction of MUCH is illustrated in~\Cref{fig:pipeline} and detailed in the following sections.

\subsection{Collecting questions} \label{sec:methodology_questions}

First, we collect questions by filtering the 200 questions in English, French, Spanish and German from Mu-SHROOM dataset~\citeplanguageresource{mushroom} (see [A] in Fig.~\ref{fig:pipeline}). We chose Mu-SHROOM because it is multilingual and well-documented, containing only factual and non-ambiguous questions. Moreover, it is designed for hallucination detection, containing difficult questions for which small LLMs are likely to hallucinate, which si necessary to obtain interesting annotations in MUCH. Finally, Mu-SHROOM provides the URL of a Wikipedia page containing the answer to each question, which is necessary for high-quality automatic annotation (see~\Cref{sec:methodology_annotation}).

\subsection{LLM generation} \label{sec:methodology_generation}

For LLM generation (see [B] in Fig.~\ref{fig:pipeline}), we use four open-weight models for generation: ``gemma-3-4b-it''~\citep{team_gemma_2025}; ``Ministral-8B-Instruct-2410''~\citep{mistral_ai_team_ministral_2025}; ``Llama-3.1-8B-Instruct''~\citep{grattafiori_llama_2024}; and ``Llama-3.2-8B-Instruct''~\citep{grattafiori_llama_2024}. We used the instruct versions to ensure the models accurately follow the prompt. Answers were generated using two temperatures (1.0 and 0.7), resulting in eight generations per question. Generations have an average length of 24.5 tokens and 97.3 characters.  Finally, we retain 24 logits per generation token, which is sufficient for most white-box methods. For example, the state-of-the-art method CCP \citep{fadeeva_fact-checking_2024} uses only 10 logits per token. Out of the 6,448 LLM generations, four contained a token that was not among the top-24 most likely tokens. For simplicity, we excluded these four samples in step [E]. The generation hyperparameters, including system prompt, user prompt, temperature, seed, library versions and other relevant settings, are provided in Appendix~\ref{app:computing} and \ref{app:generation_details}.

\subsection{Claim Segmentation} \label{sec:methodology_segmentation}

Claim segmentation is a core component of our pipeline (see [C] in Fig.~\ref{fig:pipeline}). As explained in the introduction, the segmenter needs to be fast to minimise the overhead of uncertainty quantification. It must also be reproducible and independent of any UQ method to ensure fair comparison and avoid re-annotating evaluation data for future evaluations. Finally, it must always map claims to chunks of tokens generated by the target LLM, since most white-box UQ methods rely on token-level computations. We introduce \segmenter, a new claim segmentation algorithm that meets these four requirements. It is fully rule-based and does not require external models or internet access, making it suitable for offline or computation-limited use cases. It is designed for English, French, Spanish, and German. We retain only these four European languages because their stopword and punctuation systems are similar. We expect our segmenter to be easily adaptable to languages with similar punctuation and stopwords, although we have not tested it beyond the four languages mentioned.

\paragraph{On the importance of segmentation for UQ}

The most straightforward approach to annotate the factuality of LLM generations in a fine-grained manner would be to annotate each token individually. However, individual tokens lack the contextual structure needed to represent semantic uncertainty, which arises over spans of text rather than single units. For instance, in the example of~\Cref{fig:teaser}, the non-factual part mainly involves four tokens: ``\$'', ``1'', ``,'' and ``699''. Yet, it would be arbitrary to decide whether tokens ``\$'' and ``,'' should be marked as non-factual, since they could appear in a correct response, or whether all four tokens should be marked as non-factual. This difficulty makes detecting precise span-level boundaries particularly challenging in span-level UQ benchmarks~\citep{rykov_when_2025}. Consequently, to focus on semantic hallucinations and avoid imposing token-level factuality conventions that could unnecessarily constrain future methods, we focus on claim-level factuality.

\paragraph{Our segmentation algorithm.} 

We release a PyPI implementation of our segmentation algorithm for real-world use. \Cref{alg:segmenter} presents the corresponding pseudo-code, which consists of two main steps. First, we split the LLM generation \verb|inpt| into words using an external word tokenizer, and we use these words to identify the character indices of claim starts (ll. 2-22). Second, we map these character indices to the tokens of the LLM generation (ll. 24-41). As a result, \segmenter~outputs a list of lists: each contains the token indices for a claim, for example: [[0,1], [2,3,4], [5]].

In the first step, we use \verb|nltk|'s \verb|TreebankWordTokenizer| to split \verb|inpt| into words and punctuation (l. 3), and we compare them to a list of stopwords and punctuation marks (ll. 13-14). This list contains \verb|nltk|'s stopword list in each language, plus \verb|string| punctuation list as well as additional custom punctuation marks. We avoid LLM tokenizers because they differ across models, which would make this step unreliable. When we detect a stopword or punctuation mark, we add the index of its first character to the list of claim starts, since stopwords and punctuation often introduce new ideas that form separate claims (l. 17). However, in the case of several stopwords or punctuation marks in a row, we only start one single claim (see \verb|stop_prev| in ll. 15-18). Finally, \verb|nltk| word tokenizer merges points "." to the end of the preceding word, contrary to other punctuation marks. For that reason, we introduce \verb|stop_next| to start a new claim right after a word ending with a point (ll. 15 and 19). An example of claim segmentation is provided below, with stopwords and punctuation marks in bold. We see that the three ideas in the sentence are correctly separated into three separate claims.

\[
    \overbracket{\text{\textbf{No,} Xining}}\ \overbracket{\text{\textbf{is the} largest city}}\ \overbracket{\text{\textbf{in} Qinghai.}}
\]

The second step maps the character indices to the tokens generated by the LLM (ll. 24-41). Because claims are defined by their character indices, we can always map them to LLM tokens, unlike prompt-based segmentation methods such as the one used in \citep{fadeeva_fact-checking_2024}, which fails in about 5\% of cases. Note that the end-of-sequence token always form its own claim, but we exclude this special claim in the next steps of the pipeline.

\begin{algorithm}[!t]
\includegraphics[width=\columnwidth]{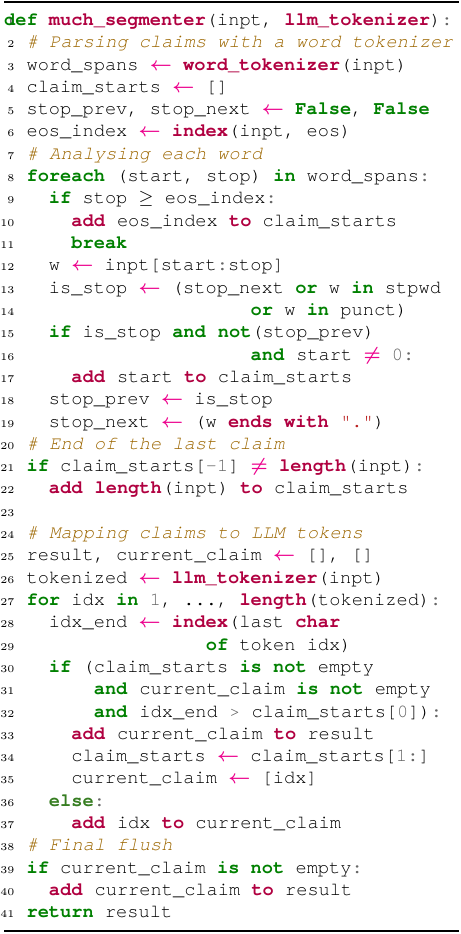}
\vspace{-.3cm}
\caption{Pseudocode of~\protect\segmenter}\label{alg:segmenter}
\end{algorithm}

\subsection{Automated Annotation} \label{sec:methodology_annotation}
\begin{figure*}[!t]
    \centering
    \definecolor{myblue}{RGB}{22,42,95}
\definecolor{myblue_dark}{RGB}{0,0,95}
\centering
\begin{subfigure}[b]{0.19\textwidth}
	\begin{tikzpicture}
		\matrix (mg) [matrix of nodes,
		nodes in empty cells,
		nodes={minimum width=3.5em, minimum height=3.5em, outer sep=0pt, anchor=center},
		column sep=-\pgflinewidth, row sep=-\pgflinewidth,
		cells={nodes={draw, font=\footnotesize}},
		] {
		|[fill=myblue!49]| \color{myblue} 10,012 & |[fill=myblue_dark!5.7]| \color{myblue} 1,162 \\
		|[fill=myblue_dark!13.2]| \color{myblue} 2,680 & |[fill=myblue]| \color{white} 20,291 \\
		};

		\node[above=.2em of $(mg-1-1.north)!0.5!(mg-1-2.north)$] {\scriptsize GPT-4.1};
		\node[left=.2em of $(mg-1-1.west)!0.5!(mg-2-1.west)$, rotate=90, anchor=south] {\scriptsize GPT-4o};

		\node[left=-.2em of mg-1-1.west] {\scriptsize -1};
		\node[left=-.2em of mg-2-1.west] {\scriptsize 1};
		\node[above=-.2em of mg-1-1.north] {\scriptsize -1};
		\node[above=-.2em of mg-1-2.north] {\scriptsize 1};
	\end{tikzpicture}
	\captionsetup{justification=centering}
	\caption{GPT-4o vs -4.1\\\textit{Before filtering}\\(\(\kappa=0.753\))}
    \label{fig:correlation_matrix_gpts}
\end{subfigure}
\begin{subfigure}[b]{0.19\textwidth}
	\begin{tikzpicture}
		\matrix (mg) [matrix of nodes,
		nodes in empty cells,
		nodes={minimum width=3.5em, minimum height=3.5em, outer sep=0pt, anchor=center},
		column sep=-\pgflinewidth, row sep=-\pgflinewidth,
		cells={nodes={draw, font=\footnotesize}},
		] {
		|[fill=myblue!42.3]| \color{myblue} 142 & |[fill=myblue_dark!4.7]| \color{myblue} 16\\
		|[fill=myblue_dark!5]| \color{myblue} 17 & |[fill=myblue]| \color{white} 336 \\
		};

		\node[above=.2em of $(mg-1-1.north)!0.5!(mg-1-2.north)$] {\scriptsize an1};
		\node[left=.2em of $(mg-1-1.west)!0.5!(mg-2-1.west)$, rotate=90, anchor=south] {\scriptsize an0};

		\node[left=-.2em of mg-1-1.west] {\scriptsize -1};
		\node[left=-.2em of mg-2-1.west] {\scriptsize 1};
		\node[above=-.2em of mg-1-1.north] {\scriptsize -1};
		\node[above=-.2em of mg-1-2.north] {\scriptsize 1};
	\end{tikzpicture}
	\captionsetup{justification=centering}
	\caption{an0 vs an1\\\textit{Before filtering}\\(\(\kappa=0.797\))}
    \label{fig:correlation_human_before}
\end{subfigure}
\vrule width 0.7pt height 4.3cm depth 0pt
\begin{subfigure}[b]{0.19\textwidth}
	\begin{tikzpicture}
		\matrix (m0) [matrix of nodes,
		nodes in empty cells,
		nodes={minimum width=3.5em, minimum height=3.5em, outer sep=0pt, anchor=center},
		column sep=-\pgflinewidth, row sep=-\pgflinewidth,
		cells={nodes={draw, font=\footnotesize}},
		] {
		|[fill=myblue!31]| \color{myblue} 192 & |[fill=myblue_dark!6.0]| \color{myblue} 37\\
		|[fill=myblue_dark!3.6]| \color{myblue} 22 & |[fill=myblue]| \color{white} 616\\
		};

		\node[above=.2em of $(m0-1-1.north)!0.5!(m0-1-2.north)$] {\scriptsize an0};
		\node[left=.2em of $(m0-1-1.west)!0.5!(m0-2-1.west)$, rotate=90, anchor=south] {\scriptsize GPT};

		\node[left=-.2em of m0-1-1.west] {\scriptsize -1};
		\node[left=-.2em of m0-2-1.west] {\scriptsize 1};
		\node[above=-.2em of m0-1-1.north] {\scriptsize -1};
		\node[above=-.2em of m0-1-2.north] {\scriptsize 1};
	\end{tikzpicture}
	\captionsetup{justification=centering}
	\caption{GPT vs an0\\\textit{After filtering}\\(\(\kappa=0.821\))}
    \label{fig:correlation_an0}
\end{subfigure}
\begin{subfigure}[b]{0.19\textwidth}
	\begin{tikzpicture}
		\matrix (m1) [matrix of nodes,
		nodes in empty cells,
		nodes={minimum width=3.5em, minimum height=3.5em, outer sep=0pt, anchor=center},
		column sep=-\pgflinewidth, row sep=-\pgflinewidth,
		cells={nodes={draw, font=\footnotesize}},
		] {
		|[fill=myblue!31.5]| \color{myblue} 196 & |[fill=myblue_dark!5.3]| \color{myblue} 33\\
		|[fill=myblue_dark!2.7]| \color{myblue} 17 & |[fill=myblue!100]| \color{white} 621\\
		};

		\node[above=.2em of $(m1-1-1.north)!0.5!(m1-1-2.north)$] {\scriptsize an1};
		\node[left=.2em of $(m1-1-1.west)!0.5!(m1-2-1.west)$, rotate=90, anchor=south] {\scriptsize GPT};

		\node[left=-.2em of m1-1-1.west] {\scriptsize -1};
		\node[left=-.2em of m1-2-1.west] {\scriptsize 1};
		\node[above=-.2em of m1-1-1.north] {\scriptsize -1};
		\node[above=-.2em of m1-1-2.north] {\scriptsize 1};
	\end{tikzpicture}
	\captionsetup{justification=centering}
	\caption{GPT vs an1\\\textit{After filtering}\\(\(\kappa=0.848\))}
    \label{fig:correlation_an1}
\end{subfigure}
\begin{subfigure}[b]{0.19\textwidth}
	\begin{tikzpicture}
		\matrix (m2) [matrix of nodes,
		nodes in empty cells,
		nodes={minimum width=3.5em, minimum height=3.5em, outer sep=0pt, anchor=center},
		column sep=-\pgflinewidth, row sep=-\pgflinewidth,
		cells={nodes={draw, font=\footnotesize}},
		] {
		|[fill=myblue!31.4]| \color{myblue} 201 & |[fill=myblue_dark!2.0]| \color{myblue} 13\\
		|[fill=myblue_dark!1.9]| \color{myblue} 12 & |[fill=myblue!100]| \color{white} 641\\
		};

		\node[above=.2em of $(m2-1-1.north)!0.5!(m2-1-2.north)$] {\scriptsize an0};
		\node[left=.2em of $(m2-1-1.west)!0.5!(m2-2-1.west)$, rotate=90, anchor=south] {\scriptsize an1};

		\node[left=-.2em of m2-1-1.west] {\scriptsize -1};
		\node[left=-.2em of m2-2-1.west] {\scriptsize 1};
		\node[above=-.2em of m2-1-1.north] {\scriptsize -1};
		\node[above=-.2em of m2-1-2.north] {\scriptsize 1};
	\end{tikzpicture}
	\captionsetup{justification=centering}
	\caption{an0 vs an1\\\textit{After filtering}\\(\(\kappa=0.922\))}
    \label{fig:correlation_humans}
\end{subfigure}
    \caption{Confusion matrices comparing claim annotations from GPT-4o, GPT-4.1, and human annotators (an0, an1), before and after sample filtering (see [E] in~\Cref{fig:pipeline} and~\Cref{sec:methodology_filtering}). Cohen’s kappa (\(\kappa\)) quantifies inter-annotator agreement. \Cref{fig:correlation_matrix_gpts}: GPT-4o vs GPT-4.1 on the 34.1k claims of the 6.4k samples before filtering. \Cref{fig:correlation_human_before}: two human annotators on the 511 claims of 100 random samples before filtering. \Cref{fig:correlation_an0}-\ref{fig:correlation_humans}: GPT vs an0 vs an1 on the 865 claims of 200 random samples after filtering.}
    \label{fig:correlation_matrix_main}
\end{figure*}

We automatically annotate the factuality of each claim (see [D] in Fig.~\ref{fig:pipeline}). Although human annotation remains the gold standard for assessing factuality, it is time-consuming and costly, particularly for large datasets. When we manually annotated a subset of the data (see~\Cref{sec:methodology_filtering}), we found that humans process roughly 350 claims per hour, which would translate to about 100 hours to annotate the full dataset. Moreover, relying on human annotation would make extending the dataset or adding new languages prohibitively expensive. Therefore, we adopt automated annotation. We discuss the quality of these annotations and compare them with human annotations in~\Cref{sec:methodology_filtering}.

Two models are used independently to provide two sets of annotations: GPT-4o (``gpt-4o-2024-11-20'') and GPT-4.1 (``gpt-4.1-2025-04-14''). Both models are prompted to classify each claim as either ``correct'' or ``incorrect''. As contextual knowledge, we provide the Wikipedia page associated with the question (see~\Cref{sec:methodology_questions}). We select three relevant chunks of approximately 120 tokens based on cosine similarity with the question, using OpenAI's ``text-embedding-3-large'' model to compute embeddings. The entire annotation process involved about 17M GPT completion tokens and 13k requests, for a total cost of roughly USD~50.

\paragraph{A binary annotation} Each claim is labeled as either factual or non-factual. In the initial stages of the project, we included a neutral label for cases where the information was ambiguous or irrelevant to the question. However, in practice, the LLM annotation rarely contained this label, and the neutral label was the root cause of most disagreements between annotators. We therefore decided to drop the neutral label and re-annotate the dataset with only positive and negative labels allowed. The filtering stage [E] (see~\Cref{sec:methodology_filtering}) replaces the neutral label by removing samples containing ambiguous claims on which GPT-4o and GPT-4.1 disagree.

\paragraph{Annotation guidelines} The complete annotation guidelines, as provided in the annotation prompt and to human annotators, are available in Appendix~\ref{app:annotation_details}. We instruct annotators to consider denial of answer as correct, since they do not introduce hallucinations. Moreover, the label ``incorrect'' should be specific to non-factual claims and not to the surrounding phrasing, as in:

\begin{align*}
    \text{\textbf{Question:} Who was the first Carolingian king?}\\
    \text{\textbf{Answer:}}\ {\color{mygreen}\underbracket{\text{\textbf{The} first Carolingian king}}_{\checkmark}}\ {\color{red!50!black}\underbracket{\text{\vphantom{g}\textbf{was} Clovis.}}_{\text{\ding{55}}}}
\end{align*}

\subsection{Filtering and Quality Verifications} \label{sec:methodology_filtering}
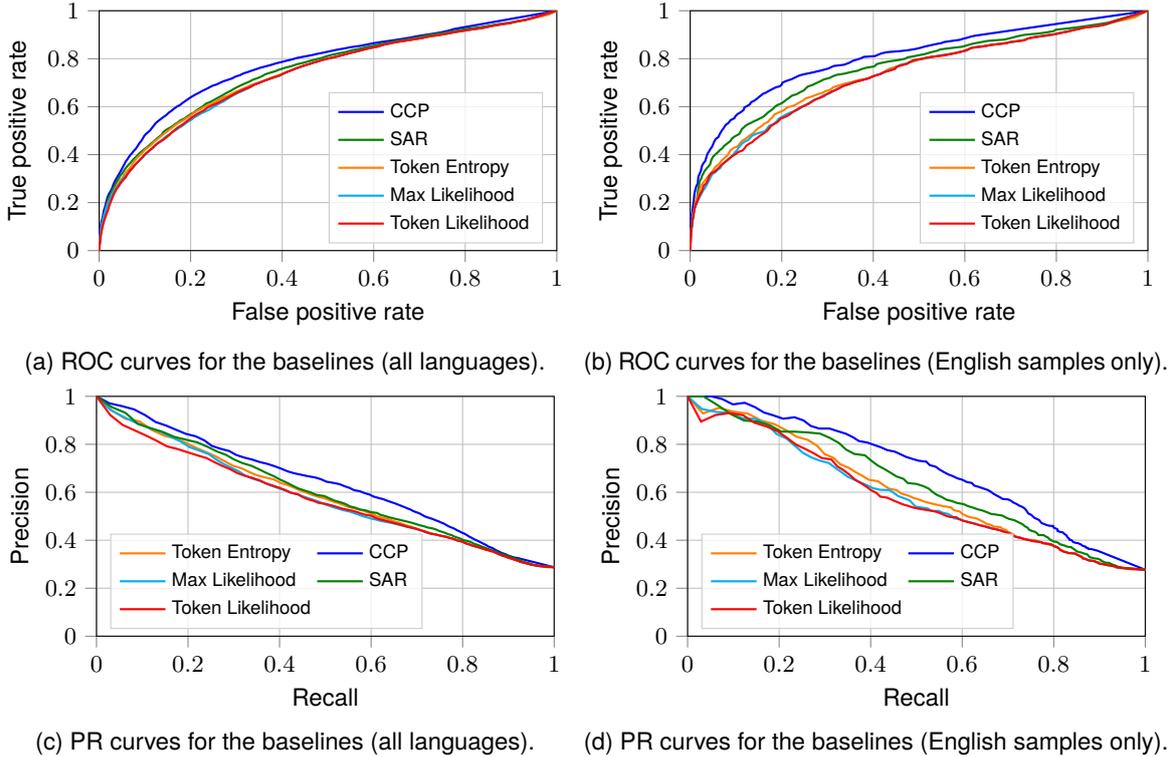
\begin{figure*}
    \centering
    \definecolor{lightgray204}{RGB}{204,204,204}
    \begin{subfigure}[t]{.48\textwidth}
        \centering
        \begin{tikzpicture}[baseline]
            \footnotesize
            \begin{axis}[
                legend pos=south east,
                legend cell align={left},
                legend style={font=\scriptsize, fill opacity=0.8, draw opacity=1, text opacity=1, draw=lightgray204},
                tick align=outside,
                tick pos=left,
                width=0.99\textwidth,
                height=.62\textwidth,
                xlabel={False positive rate},
                xmajorgrids,
                xmin=0., xmax=1., 
                ylabel={True positive rate},
                ylabel style={yshift=0pt},
                ymajorgrids,
                ymin=0., ymax=1.,
                legend columns=1,
            ]
            \addplot[blue, thick] table[y index=0, x index=1] {data/ccp_10_8_product_roc.txt};
            \addlegendentry{CCP}
            \addplot[green!50!black, thick] table[y index=0, x index=1] {data/sar_8_product_roc.txt};
            \addlegendentry{SAR}
            \addplot[orange, thick] table[y index=0, x index=1] {data/token_entropy_24_product_roc.txt};
            \addlegendentry{Token Entropy}
            \addplot[cyan, thick] table[y index=0, x index=1] {data/max_likelihood_product_roc.txt};
            \addlegendentry{Max Likelihood}
            \addplot[red, thick] table[y index=0, x index=1] {data/token_likelihood_product_roc.txt};
            \addlegendentry{Token Likelihood}
            
            \end{axis}
        \end{tikzpicture}
        \caption{ROC curves for the baselines (all languages).}
        \label{fig:roc-baselines}
    \end{subfigure}
    \begin{subfigure}[t]{.48\textwidth}
        \centering
        \begin{tikzpicture}[baseline]
            \footnotesize
            \begin{axis}[
                legend pos=south east,
                legend cell align={left},
                legend style={font=\scriptsize, fill opacity=0.8, draw opacity=1, text opacity=1, draw=lightgray204},
                tick align=outside,
                tick pos=left,
                width=0.99\textwidth,
                height=.62\textwidth,
                xlabel={False positive rate},
                xmajorgrids,
                xmin=0., xmax=1., 
                ylabel={True positive rate},
                ylabel style={yshift=0pt},
                ymajorgrids,
                ymin=0., ymax=1.,
                legend columns=1,
            ]
            \addplot[blue, thick] table[y index=0, x index=1] {data/ccp_10_8_product_en_roc.txt};
            \addlegendentry{CCP}
            \addplot[green!50!black, thick] table[y index=0, x index=1] {data/sar_8_product_en_roc.txt};
            \addlegendentry{SAR}
            \addplot[orange, thick] table[y index=0, x index=1] {data/token_entropy_24_product_en_roc.txt};
            \addlegendentry{Token Entropy}
            \addplot[cyan, thick] table[y index=0, x index=1] {data/max_likelihood_product_en_roc.txt};
            \addlegendentry{Max Likelihood}
            \addplot[red, thick] table[y index=0, x index=1] {data/token_likelihood_product_en_roc.txt};
            \addlegendentry{Token Likelihood}
        
        \end{axis}
        \end{tikzpicture}
        \caption{ROC curves for the baselines (English samples only).}
        \label{fig:roc-baselines-en}
    \end{subfigure}

    \begin{subfigure}[t]{.48\textwidth}
        \centering
        \begin{tikzpicture}[baseline]
            \footnotesize
            \begin{axis}[
                legend pos=south west,
                legend cell align={left},
                legend style={font=\scriptsize, fill opacity=0.8, draw opacity=1, text opacity=1, draw=lightgray204},
                tick align=outside,
                tick pos=left,
                width=0.99\textwidth,
                height=.62\textwidth,
                xlabel={Recall},
                xmajorgrids,
                xmin=0., xmax=1., 
                ylabel={Precision},
                ylabel style={yshift=0pt},
                ymajorgrids,
                ymin=0., ymax=1.,
                legend columns=2,
            ]
            \addplot[orange, thick] table[y index=0, x index=1] {data/token_entropy_24_product_pr.txt};
            \addlegendentry{Token Entropy}
            \addplot[blue, thick] table[y index=0, x index=1] {data/ccp_10_8_product_pr.txt};
            \addlegendentry{CCP}
            \addplot[cyan, thick] table[y index=0, x index=1] {data/max_likelihood_product_pr.txt};
            \addlegendentry{Max Likelihood}
            \addplot[green!50!black, thick] table[y index=0, x index=1] {data/sar_8_product_pr.txt};
            \addlegendentry{SAR}
            \addplot[red, thick] table[y index=0, x index=1] {data/token_likelihood_product_pr.txt};
            \addlegendentry{Token Likelihood}
            
            \end{axis}
        \end{tikzpicture}
        \caption{PR curves for the baselines (all languages).}
        \label{fig:pr-baselines}
    \end{subfigure}
    \begin{subfigure}[t]{.48\textwidth}
        \centering
        \begin{tikzpicture}[baseline]
            \footnotesize
            \begin{axis}[
                legend pos=south west,
                legend cell align={left},
                legend style={font=\scriptsize, fill opacity=0.8, draw opacity=1, text opacity=1, draw=lightgray204},
                tick align=outside,
                tick pos=left,
                width=0.99\textwidth,
                height=.62\textwidth,
                xlabel={Recall},
                xmajorgrids,
                xmin=0., xmax=1., 
                ylabel={Precision},
                ylabel style={yshift=0pt},
                ymajorgrids,
                ymin=0., ymax=1.,
                legend columns=2,
            ]
            \addplot[orange, thick] table[y index=0, x index=1] {data/token_entropy_24_product_en_pr.txt};
            \addlegendentry{Token Entropy}
            \addplot[blue, thick] table[y index=0, x index=1] {data/ccp_10_8_product_en_pr.txt};
            \addlegendentry{CCP}
            \addplot[cyan, thick] table[y index=0, x index=1] {data/max_likelihood_product_en_pr.txt};
            \addlegendentry{Max Likelihood}
            \addplot[green!50!black, thick] table[y index=0, x index=1] {data/sar_8_product_en_pr.txt};
            \addlegendentry{SAR}
            \addplot[red, thick] table[y index=0, x index=1] {data/token_likelihood_product_en_pr.txt};
            \addlegendentry{Token Likelihood}
        
        \end{axis}
        \end{tikzpicture}
        \caption{PR curves for the baselines (English samples only).}
        \label{fig:pr-baselines-en}
    \end{subfigure}
    \caption{Evaluation of baseline methods on MUCH. The state-of-the-art CCP method \citep{fadeeva_fact-checking_2024} outperforms other approaches, but there remains considerable room for improvement.}
    \label{fig:baselines}
\end{figure*}

Automated annotations are generally less reliable than human annotations. Nevertheless, recent advances in LLM technology allow them to perform many tasks with sufficiently high quality, especially when provided with a trusted knowledge base known to contain the answer (here: the Wikipedia page, see~\Cref{sec:methodology_annotation}). To ensure the quality of the data provided in MUCH, we filter out the least reliable annotations (see [E] in~\Cref{fig:pipeline}) and manually annotate a random subset of the remaining samples for quality verification (see~\Cref{fig:correlation_matrix_main}).

\paragraph{Filtering} Out of the 6,448 generations containing 34,145 claims obtained so far, we only retain 4,873 samples (75.6\% of the total) containing 20,751 claims (60.8\% of the total) in the final version of MUCH. We perform filtering by retaining only samples where GPT-4o and GPT-4.1 annotations agree on all claims. See Appendix~\ref{app:filtering_details} for details and post-filtering statistics on the MUCH benchmark. We observe that Cohen's kappa \citep{cohen_coefficient_1960} between GPT-4o and GPT-4.1 is \(\kappa=0.753\), which is close to inter-human agreement \(\kappa=0.797\) (see Figures~\ref{fig:correlation_matrix_gpts} and \ref{fig:correlation_human_before}). Such values of $\kappa \in [0.6, 0.8]$ are usually considered as a "substantial" agreement \citep{landis_measurement_1977}. However, these values indicate that many claims are ambiguous, even for humans. We remove samples containing such claims from our benchmark to avoid biasing the evaluation of future methods with ambiguous claims or unreliable annotations. On average, these ambiguous samples are longer, with 5.30 claims per sample compared to 4.26 claims per sample in MUCH.

\paragraph{Quality Verifications} After filtering, we randomly selected 50 samples per language, for a total of 867 claims. Two human annotators fluent in all four languages then independently annotated these 200 samples. They received the same instructions as those used for automated annotation (see~\Cref{sec:methodology_annotation}) and were allowed to consult any Wikipedia page. Our results appear in Figures~\ref{fig:correlation_an0}-\ref{fig:correlation_humans}. We refer to automated labels as "GPT" because GPT-4o and GPT-4.1 annotations match for these samples. We release these human annotations alongside the MUCH benchmark as a gold-standard subset.

Inter-human agreement increases substantially after filtering, confirming that ambiguous samples were effectively removed (\(\kappa = 0.922\) vs $0.797$, see Fig.~\ref{fig:correlation_humans}). The fact that humans still disagree on some claims highlights the task’s inherent subjectivity. GPT-vs-human confusion matrices are nearly diagonal, indicating strong alignment between human and automated annotators. Both annotators show similar agreement with GPT labels ($\kappa=0.821$ and $\kappa=0.848$), values usually considered  as ``almost perfect'' agreement \citep{landis_measurement_1977}. These results suggest that automated annotations are nearly as consistent with human judgments as humans are with each other. Remaining disagreements often occur when the Wikipedia page lacks the information needed to verify a claim. Overall, the automated annotations provide a reliable silver-standard benchmark for evaluating UQ methods.
\section{Hallucination statistics} \label{sec:stat_hallucination}

The proportion of sample in MUCH containing at least one non-factual claim is 60.4\% overall, but it varies considerably across models and languages, as shown in \Cref{tab:hallucination_stats}. Although Gemma-3-4b is not the smallest model in our evaluation, it shows the highest hallucination rates across all languages. The two Llama models exhibit similar overall performance, though results differ notably across languages. Ministral exhibits similar or higher values than the Llama models across all languages.

\setlength{\tabcolsep}{3pt}
\begin{table}[t]
    \begin{tabularx}{.48\textwidth}{Cccccc}
        \Xhline{.8pt}
        \noalign{\vskip .75mm}
                     & \makecell{\textbf{Llama}                                     \\ \textbf{-3.1-8B}} & \makecell{\textbf{Llama}\\ \textbf{-3.2-3B}} & \makecell{\textbf{Gemma}\\ \textbf{-3-4b}} & \makecell{\textbf{Ministral}\\ \textbf{-8B}} & \makecell{\textbf{All}}\\
        \Xhline{.8pt}
        \noalign{\vskip .75mm}
        EN           & 44.0\%                   & 38.0\% & 74.0\% & 61.7\% & 55.6\% \\
        FR           & 54.9\%                   & 48.2\% & 81.9\% & 65.0\% & 63.8\% \\
        ES           & 40.6\%                   & 51.8\% & 65.2\% & 51.2\% & 53.0\% \\
        DE           & 55.9\%                   & 55.5\% & 87.1\% & 75.6\% & 69.1\% \\
        \Xhline{.8pt}
        \noalign{\vskip .75mm}
        \textbf{All} & 48.8\%                   & 48.4\% & 76.9\% & 63.7\% & 60.4\% \\
        \Xhline{.8pt}
    \end{tabularx}
    \caption{Proportion of MUCH samples containing at least one wrong claim, per model and per language.}
    \label{tab:hallucination_stats}
\end{table}

Among samples containing at least one non-factual claim, the proportion of non-factual claims per answer is 49.8\% on average, though it varies considerably across models. It is 42.4\%, 44.2\%, and 46.6\% for Llama 3.1, Llama 3.2, and Ministral, respectively, and rises to 68.0\% for Gemma-3.

\begin{table*}[t!]
	\centering
	\begin{tabularx}{\textwidth}{lCCCCC}
		\Xhline{.8pt}
		\noalign{\vskip .75mm}
		\multirow{2}{*}{\textbf{Method}} & \multicolumn{3}{c}{\textbf{Absolute runtime (and relative to generation)}} & \multicolumn{2}{c}{\textbf{Performance}}                            \\
		                                 & Segmentation     & UQ         & Total & ROC-AUC & PR-AUC \\
		\Xhline{.8pt}
		\noalign{\vskip .75mm}
		CCP  (adapted)                   & 6s (0.2\%)       & 3,410s (124\%) & 3,416s (124\%) & 0.772   & 0.639  \\
    SAR  (adapted)                   & 6s (0.2\%)       & 613s (22.2\%)  & 619s (22.4\%) & 0.746   & 0.603  \\
		Max Likelihood                   & 6s (0.2\%)       & 8s (0.3\%)     & 14s (0.5\%)   & 0.732   & 0.582  \\
		Token Likelihood                 & 6s (0.2\%)       & 8s (0.3\%)     & 14s (0.5\%)   & 0.732   & 0.574  \\
		Token Entropy                    & 6s (0.2\%)       & 9s (0.3\%)     & 15s (0.5\%)   & 0.737   & 0.591  \\
		\Xhline{.8pt}
	\end{tabularx}
	\caption{Execution time and performance of each baseline on the MUCH benchmark. Runtime is broken down into segmentation (with~\segmenter) and UQ, and reported both in absolute terms and as a proportion of LLM generation (2,758s). We report ROC-AUC and Precision–Recall (PR) AUC.}
	\label{tab:execution_time_stats}
\end{table*}
\section{Evaluating existing baselines} \label{baselines}

\paragraph{Aggregating token-level scores into claims}

We evaluated four baselines on MUCH. Each baseline produces a token-level UQ score, which we aggregate at the claim level. In all cases, we exclude stopwords from the aggregation, following common practice \citep{fadeeva_fact-checking_2024}. We tested four aggregation strategies: the (arithmetic) mean, the maximum value, the geometric mean, and the product of token-level UQ scores. Consistent with the findings of \citet{fadeeva_fact-checking_2024}, we empirically observed that using the product yields the best performance. The results presented here rely on this aggregation strategy. Results with other aggregation strategies are available in Appendix~\ref{app:additional_results}.

\paragraph{Baselines}

We evaluated five baselines on MUCH that are frequently cited as top-performing methods. We excluded FOCUS \citep{zhang_enhancing_2023} because it relies on attention scores, which are typically unavailable in production (see discussion in~\Cref{sec:back_uq}). First, we adapted \textbf{CCP} \citep{fadeeva_fact-checking_2024}. The main adaptation arises from the fact that the meaning of claims in MUCH is often not self-contained, whereas the original method operates on self-contained sentences. To preserve the performance of the Natural Language Inference (NLI) model used in CCP, we provide the model with the $\sigma=8$ preceding and succeeding tokens of the token for which the CCP score is computed. Second, we adapted \textbf{SAR}~\citep{duan_shifting_2024}. Like CCP, it relies on an entailment score computed on sentences. For the same reason, we feed the NLI model with the $\sigma=8$ preceding and succeeding tokens around the target token (see discussion on hyperparameters in Appendix~\ref{app:baseline_details}). Third, we adapted \textbf{Maximum Likelihood} \citep{aichberger_rethinking_2024}, where the token-level UQ score corresponds to the probability of the most likely token at each generation step. Fourth, \textbf{Token Likelihood} \citep{guerreiro_looking_2023} is defined as the likelihood of the sampled token. Finally, \textbf{Token Entropy} \citep{malinin_uncertainty_2021} is the entropy of the probability distribution over the top-24 tokens provided in MUCH.

\paragraph{Evaluation metrics} For performance, we report the Area Under the ROC Curve (ROC-AUC) and the Area Under the Precision–Recall Curve (PR-AUC), which are standard metrics for claim-level UQ~\citep{vashurin_benchmarking_2025}. We also report the computation time of each method, a critical factor often omitted from benchmarks. We report computation time both in absolute terms and as a proportion of the time required for LLM generation, because UQ overhead relative to generation should remain minimal to enable large-scale, real-time uncertainty quantification. Conversely, methods such as CCP exclude segmentation from their runtime estimation, even though it adds at least a 100\% overhead due to the LLMs involved (see Introduction). The introduction of~\segmenter~in this paper substantially reduces this overhead (see~\Cref{tab:execution_time_stats}).

\paragraph{A poor performance–efficiency trade-off} Evaluation results are shown in~\Cref{fig:baselines} and~\Cref{tab:execution_time_stats}. Existing baselines achieve reasonable performance, with CCP reaching a ROC-AUC of 0.772. This value is close to the ROC-AUC of 0.74 reported for CCP on English/Mistral 7B in the private test of \citet{vashurin_benchmarking_2025}, indicating that our adaptation preserved the baseline’s original behavior. The other baselines achieve slightly lower ROC-AUC scores. However, CCP is substantially more computationally expensive, with an overhead of about 124\% of LLM generation time, for a modest gain in UQ performance.

\paragraph{Discussion} Our evaluation shows that existing claim-level UQ methods still have substantial room for improvement. First, realistic applications require a low false positive rate and high precision. However, the best method evaluated here achieves only a TPR@FPR=10\% of 48\% and a Rec@Prec=80\% of 23.5\%, which remains too low for reliable UQ. We encourage future methods to focus on improving performance in these regions of the ROC and PR curves. Second, current methods perform better in English than in other languages (see~\Cref{fig:roc-baselines-en}) because they rely on NLI models that are more performant in English. Future methods should aim to close the gap in UQ performance across languages. Third, future methods should report computation time relative to LLM generation and aim for minimal overhead, on the order of a few percent, to enable large-scale adoption for real-time monitoring of LLM outputs. To promote transparency regarding computation time and comparison with LLM generation runtime, we release a script to estimate the generation runtime on a given machine.

\section*{Conclusion} \label{conclusion}

We release MUCH, a multilingual claim-level uncertainty quantification benchmark comprising 4.8k samples representing 20.7k claims. The benchmark includes factuality labels and 24 logits per token, supporting the development of new white-box methods  and their fair and reproducible evaluation.

We also advocate for the use of a reproducible and computation-efficient segmentation algorithm, independent of any uncertainty quantification method. To this end, we introduce~\segmenter, a claim segmentation package for English, French, Spanish, and German, available on PyPI under a permissive license.

Finally, we benchmark existing strong baselines for claim-level uncertainty quantification. Our results show substantial room for improvement in both accuracy and efficiency. For performance, future methods should target low-FPR/high-precision regimes to support realistic deployment. For efficiency, future method should report runtime relative to LLM generation, aiming for minimal overhead to enable real-time monitoring of LLM outputs.

\section*{Limitations} \label{sec:limitations}

\paragraph{Limited number of languages} Our benchmark includes only four European languages: English, French, Spanish, and German. We focus on these languages because their punctuation and stopword systems are very similar, enabling the use of a single claim segmentation algorithm without requiring language detection. This choice simplifies and accelerates the segmentation. While these four languages cover many practical scenarios, expanding the benchmark to include additional languages would improve its generality. In particular, incorporating languages that do not use the Latin alphabet, such as Chinese, Arabic, or Russian, would be valuable, though it would increase the complexity of the segmenter.

\paragraph{Automated annotation vs human annotations} During manual annotations performed for quality verification, we observed that, for some samples, assessing the factuality of an LLM generation requires consulting pages other than the one linked to the question in Mu-SHROOM~\citeplanguageresource{mushroom}. This typically occurs when the LLM produces an unexpected yet potentially factual answer whose verification is nontrivial. Incorporating a human-generated \textit{gold answer} for each question, as in \citep{rykov_when_2025}, could help improve the quality of automated annotations. Finally, allowing the annotation model to access multiple Wikipedia pages through a multi-agent pipeline with web search could further enhance annotation quality, albeit at a substantial computational cost.

\paragraph{Unavailability of attention weights} We do not provide the attention weights of samples in MUCH, making our benchmark unsuitable for evaluating attention-based UQ methods such as FOCUS~\citep{zhang_enhancing_2023}. To mimic realistic LLM generation conditions, the MUCH samples were generated with the FlashAttention~\citep{dao_flashattention_2022, dao_flashattention-2_2023} algorithm, which does not allow retrieving attention weights. Because such implementations are now widely adopted, attention-based UQ methods are inapplicable in most scenarios.

\section*{Code and Data}

Alongside this paper, we provide: the MUCH dataset (including LLM generations, logits, annotations, and generation configurations); the \segmenter~source code and PyPI package; token-level scores of all baselines evaluated; and a repository containing the source code, seeds, library versions, and hyperparameters necessary to reproduce the generation of MUCH and the evaluation of the baselines, along with a script to estimate LLM generation time on any machine.

\begin{center}
  \begin{minipage}{0.45\linewidth}
    \centering
    \raisebox{-0.2\height}{\includegraphics[width=1em]{data/logos/github-mark.png}}%
    \hspace{0.5em}%
    {\small\texttt{\href{https://github.com/orailix/much}{\tt orailix/much}}}
  \end{minipage}
  \begin{minipage}{0.45\linewidth}
    \centering
    \raisebox{-0.2\height}{\includegraphics[width=1em]{data/logos/huggingface_logo-noborder.pdf}}%
    \hspace{0.5em}%
    {\small\texttt{\href{https://huggingface.co/datasets/orailix/MUCH}{\tt orailix/MUCH}}}
  \end{minipage}
  \begin{minipage}{0.45\linewidth}
    \centering
    \raisebox{-0.2\height}{\includegraphics[width=1em]{data/logos/pypi_logo.png}}%
    \hspace{0.5em}%
    {\small\texttt{\href{https://pypi.org/project/much-segmenter/}{\tt much-segmenter}}}
  \end{minipage}
\end{center}

\section*{Acknowledgement}

As detailed in~\Cref{sec:methodology}, the construction of MUCH benchmark involves some fields extracted from Mu-SHROOM dataset~\citeplanguageresource{mushroom}, a dataset released under CC-BY-4.0 license.

We thank Mahammed El Sharkawy, Lucas Thil, Mohamed Dhouib, Clément Elliker, Mathis Le Bail, Benoit Goupil and Martin Bonsergent-Brachet for discussions on early versions of this paper.

This work received financial support from the research chair \textit{Trustworthy and Responsible AI} at École Polytechnique.  

This work was granted access to the HPC resources of IDRIS under the allocation AD011014843R1 made by GENCI.

\section*{Bibliographical References}\label{sec:reference}

\bibliographystyle{lrec2026-natbib}
\bibliography{references}

\section*{Language Resource References}
\label{lr:ref}
\bibliographystylelanguageresource{lrec2026-natbib}
\bibliographylanguageresource{languageresource}

\appendix

\section{Computing infrastructure} \label{app:computing}

Experiments were conducted on two devices: an HPC cluster and a local machine. The HPC cluster comprises nodes equipped with 2× AMD EPYC 7543 CPUs (32 cores, 64 threads each), 468 GB of RAM, and 8× NVIDIA A100 GPUs (80 GB each), running Red Hat Enterprise Linux~9.4. The computations on the cluster were deployed on jobs allocated 1/8 of a node (1~GPU, 8~cores, and 58.5~GB of RAM). The local machine is equipped with an Apple M4 Pro and 48~GB of RAM, running MacOS~15.6.1. Relevant software versions: Python~3.12.0, \verb|accelerate|~1.17.0, \verb|datasets|~3.6.0, \verb|numpy|~2.3.0, \verb|torch|~2.7.1, \verb|transformers|~4.52.4.

The main computation steps were divided as described in~\Cref{tab:computation_separation}. The HPC cluster was used for GPU-intensive computations (LLM generations, as well as CCP and SAR baselines due to the use of an NLI model).

\begin{table}[h!]
    \begin{tabularx}{.48\textwidth}{Cc}
        \Xhline{.8pt}
        \noalign{\vskip .75mm}
        \textbf{Step} & \textbf{Machine} \\
        \Xhline{.8pt}
        \noalign{\vskip .75mm}
        Question generation & Local \\
        LLM generation & HPC cluster \\
        Segmentation & Local \\
        Wikipedia page caching & Local \\
        GPT annotation & Local \\
        Baselines: TL, MxL, TE & Local \\
        Baselines: CCP and SAR & HPC cluster \\
        Baseline evaluation & Local \\
        \Xhline{.8pt}
    \end{tabularx}
    \caption{Separation of computation steps between HPC cluster and local machine.}
    \label{tab:computation_separation}
\end{table}

\section{Generation details} \label{app:generation_details}

The hyperparameters used for LLM generation of MUCH samples are provided in~\Cref{tab:hyperparameters}. The system and user prompts used for generation are shown in~\Cref{fig:generation_prompt}. Generating the 6,448 samples took 4,540~seconds. Since only 4,873 samples are retained in MUCH and these samples contain an average of 4.26 claims, compared to 5.30 for all samples, we estimate the LLM generation time for MUCH as $4\text{,}540 \times (4\text{,}873/6\text{,}448) \times (4.26/5.30) = 2\text{,}758$~seconds.

\section{Annotation instructions} \label{app:annotation_details}

The instructions provided to human annotators and the GPT annotation pipeline are detailed in~\Cref{fig:annotation_instructions}. These instructions were concatenated with the reference knowledge extracted from the Wikipedia page associated with each question. The reference knowledge consists of (i) the three chunks of approximately 120 tokens from the page that exhibit the highest cosine similarity with the question, and (ii) the \textit{infobox} of the page. The \textit{infobox} is the fixed-format table typically displayed in the top right-hand corner of articles to present a consistent summary of relevant information.

\section{Baseline Details} \label{app:baseline_details}

\paragraph{NLI model for SAR and CCP} SAR and CCP rely on an NLI model. We used the same model for both baselines: \verb|microsoft/deberta-large-mnli| \citep{he_deberta_2021}. This choice was motivated by its use in the original implementation of CCP~\citep{fadeeva_fact-checking_2024}. The main limitation of this model is that it is not optimized for French, Spanish, and German, which is particularly problematic for Spanish, where CCP and SAR achieved poor results (see Appendix~\ref{app:additional_results}).

\paragraph{Hyperparameters of CCP, SAR and Token Entropy} We evaluated SAR with three hyperparameters: $\sigma \in \{3, 5, 8\}$. We recall that $\sigma$ refers to the number of preceding and succeeding tokens passed as context to the NLI model. In the main part of the paper, “SAR” refers to the results with $\sigma = 8$, as this configuration achieves the best performance while maintaining an acceptable execution time (see~\Cref{tab:appendix_execution_and_perf}). On the other hand, CCP relies on two hyperparameters. The first one is $\sigma$, which plays the same role as in SAR. The second metric is the number of possibilities evaluated per token, denoted by $\delta$. For instance, since we only provide 24 logits per generated token, the maximum value of $\delta$ compatible with MUCH is 24. In the original paper, \citet{fadeeva_fact-checking_2024} used $\delta=10$. We evaluated the CCP baseline with six hyperparameters: $\delta \in \{10, 24\}$ and $\sigma \in \{3, 5, 8\}$. In the main part of the paper, “CCP” refers to the results with $\delta = 10$ and $\sigma = 8$, since this configuration provides near-optimal performance while keeping computation time lower than the other CCP settings (see~\Cref{tab:appendix_execution_and_perf}). Moreover, the choice of $\delta = 10$ is consistent with the original paper~\citep{fadeeva_fact-checking_2024}, and the choice of $\sigma = 8$ is consistent with that made for the SAR baseline. Finally, Token Entropy depends on a single hyperparameter: the number of possibilities considered when computing the entropy of the distribution for each generated token. We denote it by $\delta$ and evaluate three values, $\delta \in \{5, 10, 24\}$. In the main part of the paper, we selected $\delta=24$ because it obtains the best results with a negligible increase of the computational cost. The performance of each configuration is discussed in~\Cref{app:additional_results}.

\section{Samples filtered out for quality reasons} \label{app:filtering_details}

Out of the 6,448 generated samples, we retain only 4,873 in MUCH. Samples were filtered out for three reasons:

\begin{itemize}
    \item 1,568 samples were removed because the labels annotated by GPT-4o and GPT-4.1 mismatch on at least one claim;
    \item 4 samples were removed because one of their tokens was sampled outside the top-24 most likely tokens;
    \item 3 samples were removed because they did not include an EOS token. These cases correspond to generation loops where the model endlessly repeats the same text snippet.
\end{itemize}

The proportion of samples per language and per model after the filtering step is presented in~\Cref{tab:lang_and_models_stats}.
\setlength{\tabcolsep}{3pt}
\begin{table}[h!]
    \begin{tabularx}{.48\textwidth}{@{}Cccccc@{}}
        \Xhline{.8pt}
        \noalign{\vskip .75mm}
                     & \makecell{\textbf{Llama}                                     \\ \textbf{-3.1-8B}} & \makecell{\textbf{Llama}\\ \textbf{-3.2-3B}} & \makecell{\textbf{Gemma}\\ \textbf{-3-4b}} & \makecell{\textbf{Ministral}\\ \textbf{-8B}} & \makecell{\textbf{Sum}}\\
        \Xhline{.8pt}
        \noalign{\vskip .75mm}
        EN           & 6.1\%                   & 6.2\% & 7.3\% & 7.2\% & 26.8\% \\
        FR           & 5.2\%                   & 5.2\% & 6.7\% & 6.1\% & 23.3\% \\
        ES           & 5.5\%                   & 5.7\% & 7.0\% & 5.9\% & 24.0\% \\
        DE           & 6.0\%                   & 6.5\% & 6.9\% & 6.6\% & 26.0\% \\
        \Xhline{.8pt}
        \noalign{\vskip .75mm}
        \textbf{Sum} & 22.8\%                   & 23.6\% & 27.8\% & 25.8\% & 100\% \\
        \Xhline{.8pt}
    \end{tabularx}
    \caption{Proportion samples per language and per model in MUCH benchmark.}
    \label{tab:lang_and_models_stats}
\end{table}

\section{Additional Results} \label{app:additional_results}

\begin{figure*}[!t]
    \centering
    \includegraphics[width=\textwidth]{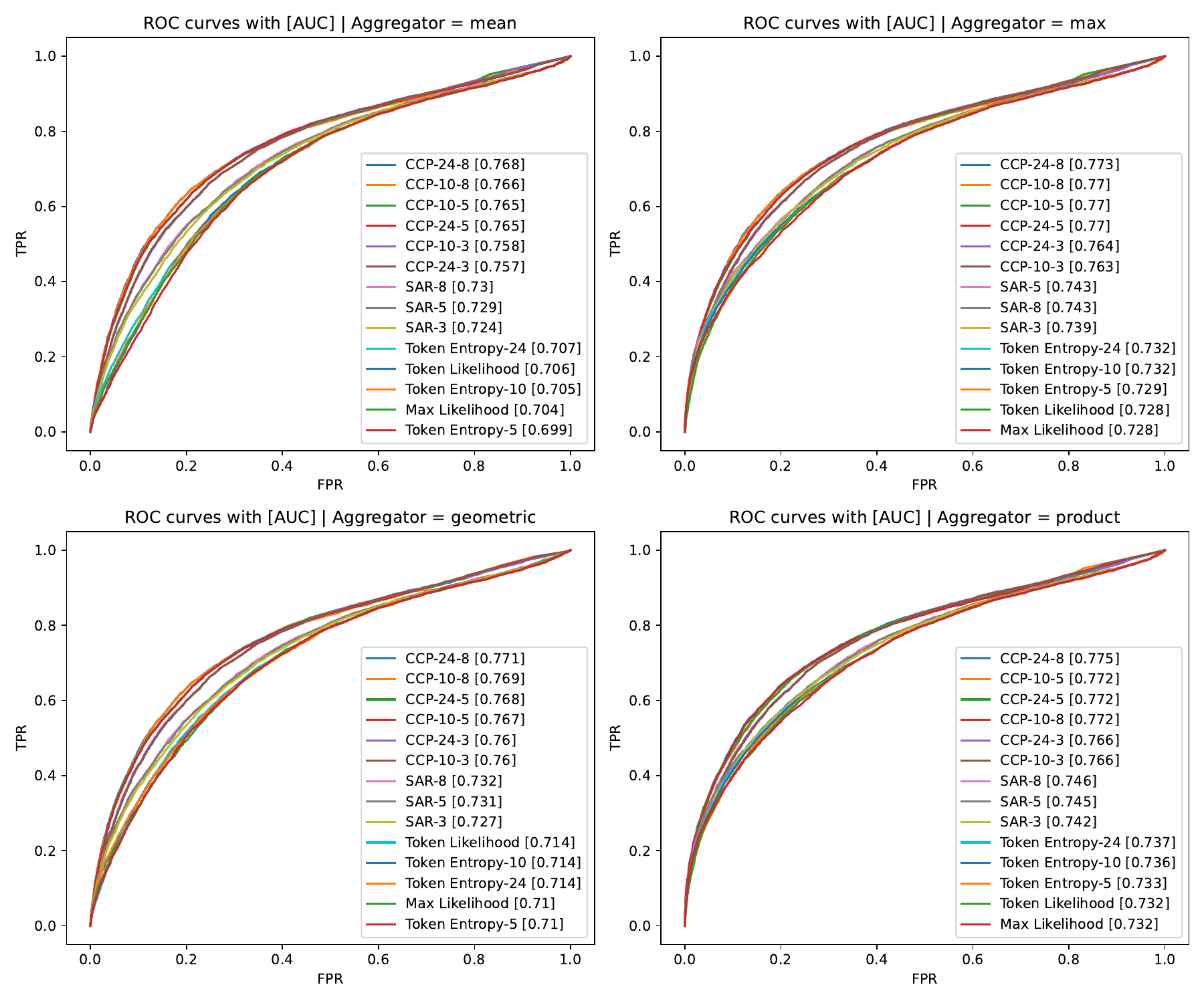}
    \caption{ROC curves for all baselines, depending on the aggregator used for evaluation.}
    \label{fig:roc_all_lang}
\end{figure*}

\begin{figure*}[!t]
    \centering
    \includegraphics[width=\textwidth]{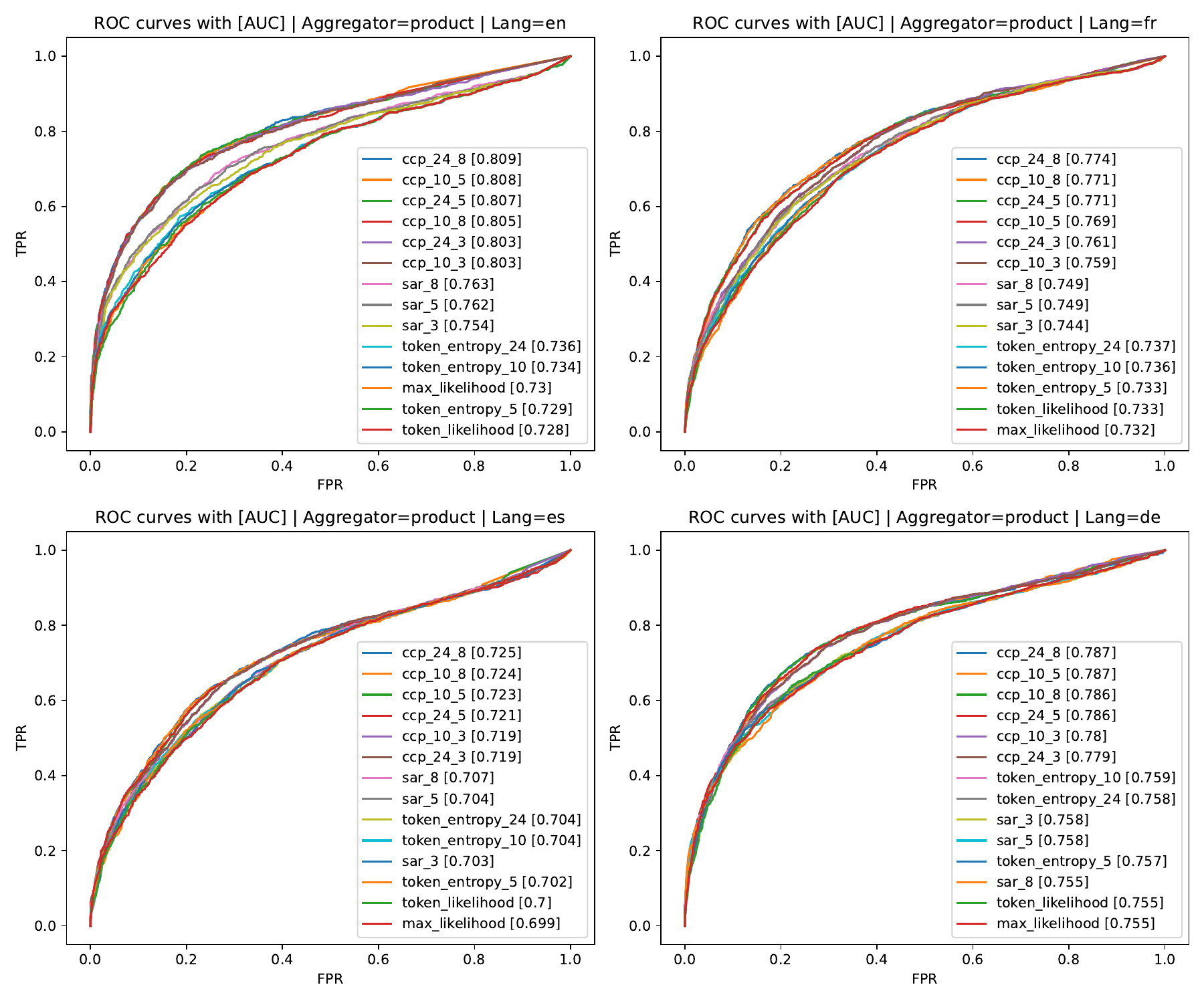}
    \caption{ROC curves for all baselines, grouped by language.}
    \label{fig:roc_per_lang}
\end{figure*}

\begin{figure*}[!t]
    \centering
    \includegraphics[width=\textwidth]{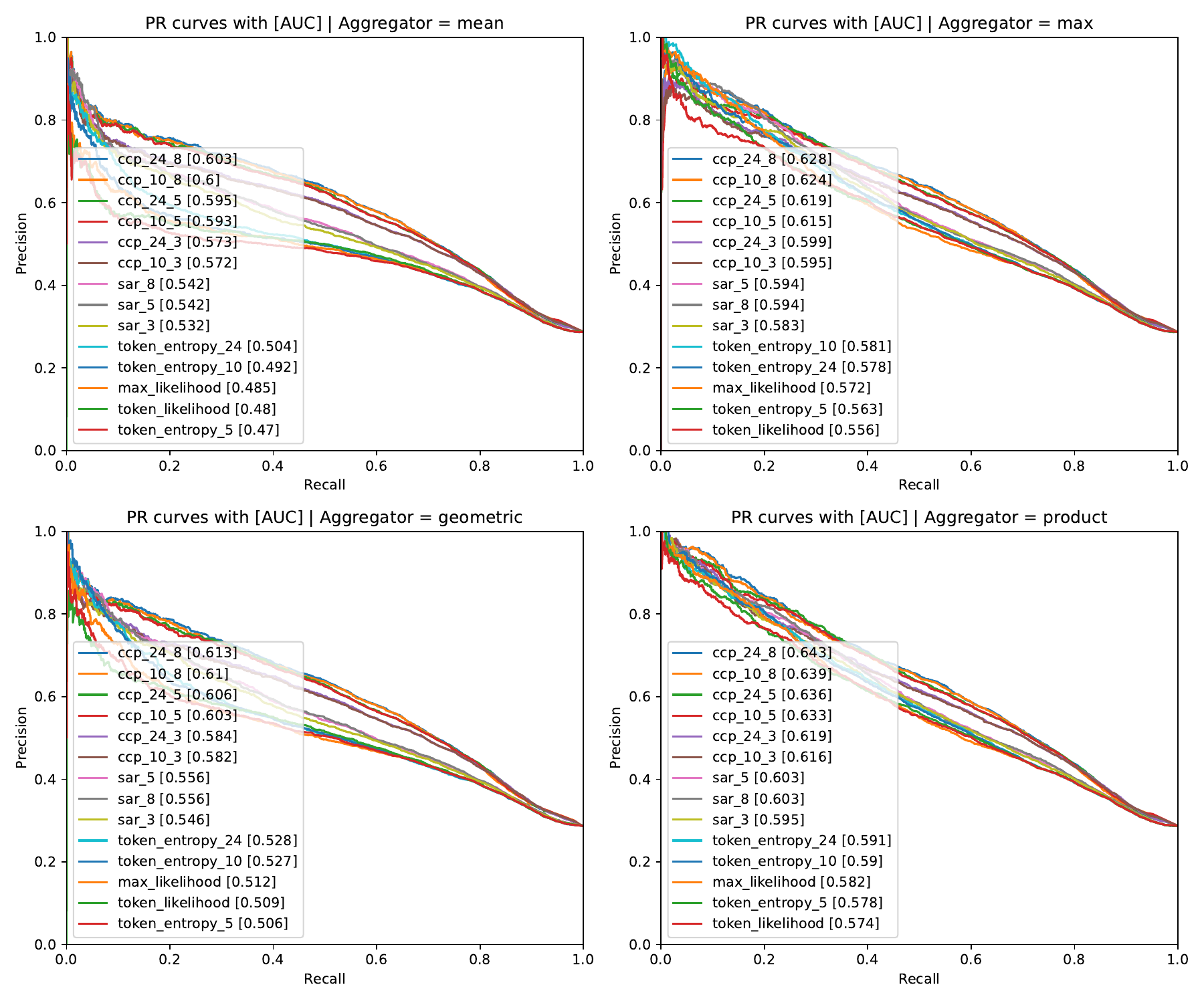}
    \caption{Precision-Recall curves for all baselines, depending on the aggregator used for evaluation.}
    \label{fig:pr_all_lang}
\end{figure*}

\begin{figure*}[!t]
    \centering
    \includegraphics[width=\textwidth]{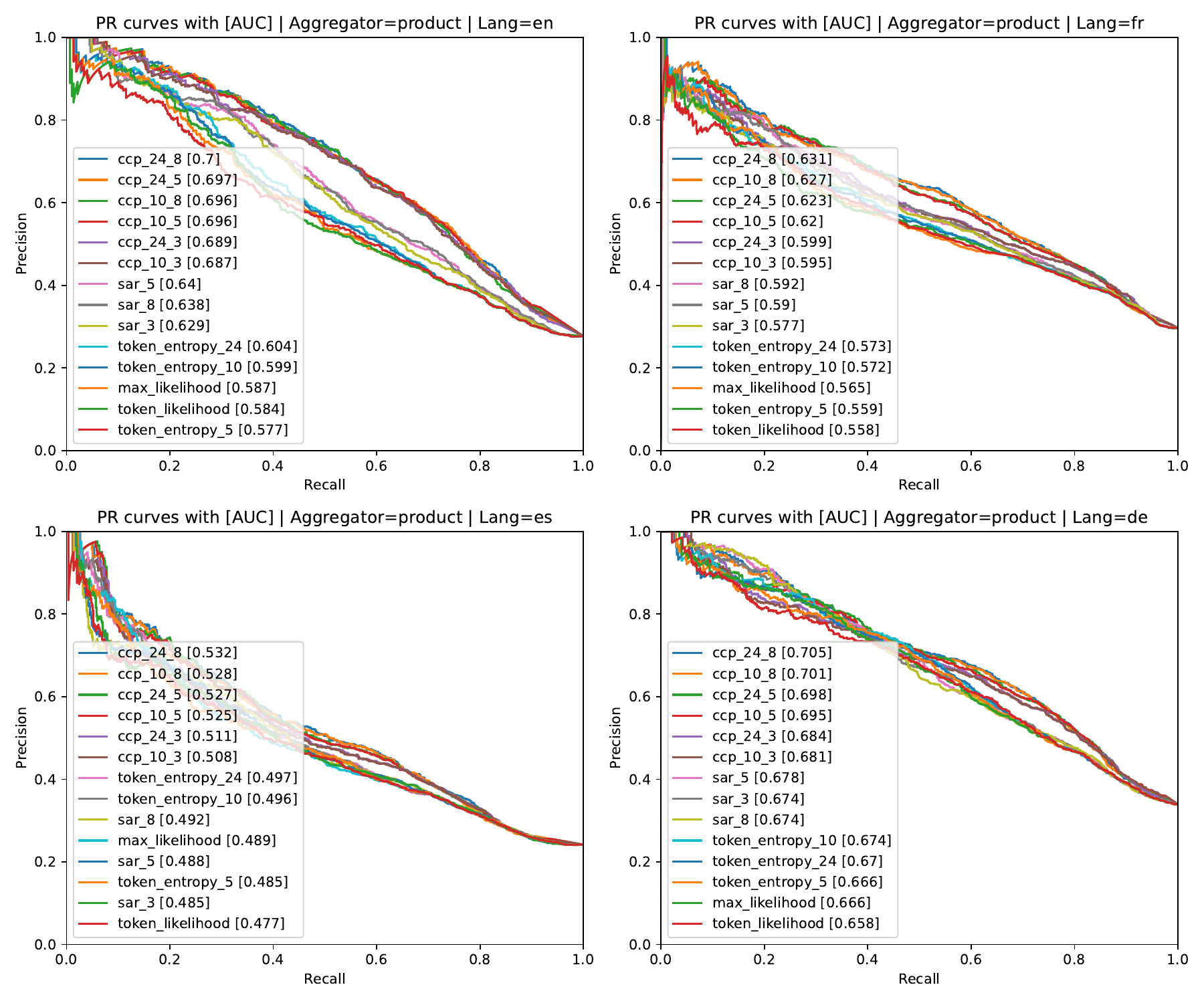}
    \caption{Precision-Recall curves for all baselines, grouped by language.}
    \label{fig:pr_per_lang}
\end{figure*}

\Cref{tab:appendix_execution_and_perf} extends~\Cref{tab:execution_time_stats} with the performance for all hyperparameter configurations of SAR, CCP, and Token Entropy. The numbers appended to the baseline names refer to the hyperparameters. For CCP, the first number corresponds to $\delta$ and the second to $\sigma$. For SAR, the number corresponds to $\sigma$, and for Token Entropy, it corresponds to $\delta$. Figure~\ref{fig:roc_all_lang} (resp.~\ref{fig:pr_all_lang}) display the ROC curve (resp. Precision–Recall curve) for all baselines, depending on the aggregator used for evaluation. The aggregator refers to the algorithm used to combine token-level UQ scores into a claim-level UQ value. Following~\citep{fadeeva_fact-checking_2024}, we evaluated four aggregators: the (arithmetic) mean of token values in the claim, the maximum, the geometric mean, and the product. Consistent with the observations of~\citep{fadeeva_fact-checking_2024}, our best results are obtained when using the product as aggregator. Figures~\ref{fig:roc_per_lang} and~\ref{fig:pr_per_lang} display the ROC and Precision–Recall curves for all baselines grouped by language. We observe that SAR and CCP achieve poor performance for languages other than English, and especially in Spanish. This can be explained by the fact that the NLI model used for these baselines was only optimized for English (see Appendix~\ref{app:baseline_details}). Evaluating computation time and performance across languages for alternative NLI models would be a promising direction for improving these baselines.

\begin{table*}[t!]
	\centering
	\begin{tabularx}{\textwidth}{ccC}
		\Xhline{.8pt}
        \noalign{\vskip .75mm}
        \textbf{Parameter} & \textbf{Type} & \textbf{Value(s)} \\
        \Xhline{.8pt}
        \noalign{\vskip .75mm}
        Model Name & Variable & \makecell[c]{\texttt{meta-llama/Llama-3.2-3B-Instruct} \\ \texttt{meta-llama/Llama-3.1-8B-Instruct} \\ \texttt{mistralai/Ministral-8B-Instruct-2410} \\ \texttt{google/gemma-3-4b-it}} \\
        \Xhline{.8pt}
        \noalign{\vskip .75mm}
        Temperature & Variable & \texttt{\{1.0, 0.7\}} \\
        \Xhline{.8pt}
        \noalign{\vskip .75mm}
        Eager & Fixed & \texttt{False} \\
        Greedy & Fixed & \texttt{False} \\
        Seed & Fixed & \texttt{1234} \\
        Top p & Fixed & \texttt{0.9} \\
        Top k & Fixed & \texttt{20} \\
        Max new tokens & Fixed & \texttt{500} \\
        \Xhline{.8pt}
        \noalign{\vskip .75mm}
    \end{tabularx}
    \caption{Hyperparameters of the LLM generations used for MUCH benchmark. We used two variable hyperparameter with 4 and 2 possibilities, leading to a total of 8 different configurations.}
    \label{tab:hyperparameters}
\end{table*}

\begin{figure*}[t!]
    \begin{tcolorbox}[colback=green!5,colframe=green!40!black,width=\textwidth,title=Annotation Instructions given to both human annotators and GPT annotation pipeline]
    
    \begin{center}\textbf{Task}\end{center}
    
    Given the original \textbf{question}, the model's \textbf{answer}, a \textbf{segmented version} of that
    answer, and some \textbf{reference knowledge}, assign a factuality score to each segment.
    
    \vspace{5pt}  \begin{center}\textbf{Scoring Convention}\end{center}
    
    -1 → Factually incorrect / contradicts previous ideas / unverifiable / approximative \\
    1 → Factually correct
    
    Be \textbf{tough and rigorous}: Only assign "1" when the content is clearly factually correct
    and relevant to the question.
    
    \begin{center}\textbf{Edge Cases}\end{center}
    
    1. \textbf{Denial of answer:} \\
    If the model explicitly states it cannot answer or denies having information
    (e.g., "I did not find any information..."), assign label 1.
    The denial itself is factually correct.
    
    \vspace{5pt} 2. \textbf{Targeted labels:} \\
    When a segment is incorrect, apply the label to the specific factual claim, not the surrounding phrasing.
    Example: Question = "Who was the first Carolingian king?" \\
    Answer = "The first Carolingian king was Clovis." \\
    Here, the incorrect part is "Clovis," not the framing "The first Carolingian...".
    
    \vspace{5pt} 3. \textbf{Numbers:} \\
       3a. \textit{Exact values required} (e.g., dates, rank in a competition, small quantities <100): \\
            - Exact match → 1 \\
            - Otherwise → -1
            
    \vspace{5pt} 3b. \textit{Approximate values acceptable} (e.g., population, area): \\
            - Exact match → 1 \\
            - Within 10\% error with "around", "approximately" or similar context → 1 \\
            - Within 10\% error without such context → -1 \\
            - Larger error → -1
    
    \vspace{5pt} 4. \textbf{Chunk granularity:} \\
    If a chunk contains any factually false statement, the entire chunk should be labeled -1.
    
    \vspace{5pt} 5. \textbf{Overgeneration mistakes and false details:} \\
    If the answer contains overgeneration that adds details to an incorrect claim (e.g., stating the citizenship
    or birthdate of the wrong person, or providing details about an event that did not occur), assign -1 to both
    the incorrect claim and any overgenerated or unverifiable details associated with it.
    
    \vspace{5pt} 6. \textbf{EOS token} \\
    End-of-sequence tokens such as \verb|<eos>|, \verb|</s>| or \verb!<|eot_id|>! should be labeled 1, except if they appear
    in the same chunk as incorrect information (see rule 4).
    
    \end{tcolorbox}
    \caption{Annotation instructions}
    \label{fig:annotation_instructions}
\end{figure*}

\begin{figure*}[t!]
    \begin{tcolorbox}[colback=blue!5,colframe=blue!40!black,width=\textwidth,title=Prompt used for the generation of MUCH dataset]
    \textbf{System:} You are a helpful assistant. Always answer questions directly. Your answers should be very concise and precise. \\
    \textbf{User:} \verb|<question>| \\
    \textbf{Assistant:} 
    \end{tcolorbox}
    \caption{Generation prompt}
    \label{fig:generation_prompt}
\end{figure*}

\begin{table*}[t!]
	\centering
	\begin{tabularx}{\textwidth}{lCCCCC}
		\Xhline{.8pt}
		\noalign{\vskip .75mm}
		\multirow{2}{*}{\textbf{Method}} & \multicolumn{3}{c}{\textbf{Absolute runtime (and relative to generation)}} & \multicolumn{2}{c}{\textbf{Performance}}                            \\
		                                 & Segmentation     & UQ         & Total & ROC-AUC & PR-AUC \\
		\Xhline{.8pt}
		\noalign{\vskip .75mm}
		CCP-24-8          & 6s (0.2\%)       & 5,429s (197\%) & 5,435s (197\%) & 0.775   & 0.643  \\
    CCP-10-5          & 6s (0.2\%)       & 3,230s (117\%) & 3,236s (117\%) & 0.772   & 0.633  \\
    CCP-24-5          & 6s (0.2\%)       & 4,268s (155\%) & 4,274s (155\%) & 0.772   & 0.636  \\
    CCP-10-8          & 6s (0.2\%)       & 3,410s (124\%) & 3,416s (124\%) & 0.772   & 0.639  \\
    CCP-24-3          & 6s (0.2\%)       & 3,508s (127\%) & 3,514s (127\%) & 0.766   & 0.619  \\
    CCP-10-3          & 6s (0.2\%)       & 4,047s (147\%) & 4,053s (147\%) & 0.766   & 0.616  \\
    SAR-8             & 6s (0.2\%)       & 613s (22.2\%)  & 619s (22.4\%)  & 0.746   & 0.603  \\
    SAR-5             & 6s (0.2\%)       & 510s (18.5\%)  & 516s (18.7\%)  & 0.745   & 0.603  \\
    SAR-3             & 6s (0.2\%)       & 419s (15.2\%)  & 425s (15.5\%)  & 0.742   & 0.595  \\
    Token Entropy-24  & 6s (0.2\%)       & 9s (0.3\%)     & 15s (0.5\%)    & 0.737   & 0.591  \\
    Token Entropy-10  & 6s (0.2\%)       & 9s (0.3\%)     & 15s (0.5\%)    & 0.736   & 0.590  \\
    Token Entropy-5   & 6s (0.2\%)       & 9s (0.3\%)     & 15s (0.5\%)    & 0.733   & 0.578  \\
		Max Likelihood    & 6s (0.2\%)       & 8s (0.3\%)     & 14s (0.5\%)    & 0.732   & 0.582  \\
		Token Likelihood  & 6s (0.2\%)       & 8s (0.3\%)     & 14s (0.5\%)    & 0.732   & 0.574  \\
		\Xhline{.8pt}
	\end{tabularx}
	\caption{Execution time and performance of each baseline on the MUCH benchmark. Runtime is broken down into segmentation (with~\segmenter) and UQ, and reported both in absolute terms and as a proportion of LLM generation (2,758s). We report ROC-AUC and Precision–Recall (PR) AUC. The numbers appended to the baseline names refer to the hyperparameters. For CCP, the first number corresponds to $\delta$ and the second to $\sigma$. For SAR, the number corresponds to $\sigma$, and for Token Entropy, it corresponds to $\delta$.}
	\label{tab:appendix_execution_and_perf}
\end{table*}

\end{document}